# Analyzing FOMC Minutes: Accuracy and Constraints of Language Models

*Wonseong Kim[1]\*, Jan Frederic Spörer[2], Siegfried Handschuh[3]*

[1] *University of St. Gallen, Institute of Computer Science (Affiliated in Korea University), e-mail: wonseongkim@korea.ac.kr*
[2] *University of St. Gallen, Institute of Computer Science, e-mail: jan.spoerer@unisg.ch*
[3] *University of St. Gallen, Institute of Computer Science, e-mail: siegfried.handschuh@unisg.ch*
\* *Corresponding Author*

## Abstract

This research article analyzes the language used in the official statements released by the Federal Open Market Committee (FOMC) after its scheduled meetings to gain insights into the impact of FOMC official statements on financial markets and economic forecasting. The study reveals that the FOMC is careful to avoid expressing emotion in their sentences and follows a set of templates to cover economic situations. The analysis employs advanced language modeling techniques such as VADER and FinBERT, and a trial test with GPT-4. The results show that FinBERT outperforms other techniques in predicting negative sentiment accurately. However, the study also highlights the challenges and limitations of using current NLP techniques to analyze FOMC texts and suggests the potential for enhancing language models and exploring alternative approaches.

# 1. Introduction

The Federal Open Market Committee (FOMC) plays a pivotal role in shaping the trajectory of the United States' monetary policy and, by extension, the global economy. As the primary decision-making body of the Federal Reserve System, the FOMC's actions and communication strategies have a profound impact on financial markets, investment decisions, and overall economic stability. The official statements released by the FOMC after each of its eight scheduled meetings per year serve as a potent force in the global economy, akin to a water drop whose ripples spread across the world, altering economic agents' expectations, and driving market fluctuations. Given the significance of these statements, it is essential to understand their influence on financial markets and economic agents. This study aims to elucidate the importance of FOMC official statements by examining their content, market reactions, and implications for monetary policy and economic forecasting.

In recent years, central banks around the world have placed greater emphasis on transparency and communication as essential tools for effective monetary policy. The FOMC's official statements are a prime example of this communication strategy, serving as a critical channel for conveying the Committee's policy stance and forward guidance. While the content of these statements has evolved over time, their relevance and impact on financial markets remain undiminished. Market participants, policymakers, and researchers alike scrutinize the language and tone of FOMC statements for signals that may guide their decision-making and forecasting processes.

This research article offers a unique technical contribution to the ongoing academic debate surrounding the role of central bank communication in influencing financial markets and the economy. While numerous studies have explored FOMC communication, few have delved deeply into the linguistic aspects of the FOMC communication text itself. Our work serves as a milestone in this domain, focusing on a linguistic approach to better understand these crucial documents. We employ a multi-method approach, integrating qualitative content analysis and quantitative techniques to investigate the impact of FOMC official statements. Our analysis encompasses the period from the inception of the FOMC's meeting minutes on 2006/01/03 to 2023/02/22, offering a comprehensive examination of these critical communication tools.

The remainder of this article is organized as follows: Section 2 provides a review of the existing literature on FOMC communication. Section 3 describes our data, including the data sources and findings from Natural

Language Processing (NLP) techniques. Section 4 presents our methods, highlighting the prediction of sentiment in FOMC meeting minutes. Finally, Sections 5 and 6 present the predicted results along with a discussion of the study's implications, limitations, and avenues for future research, emphasizing the power of FOMC statements as catalysts for economic waves that shape the global financial landscape.

## 2. Related Work

Language and communication have been increasingly recognized as important factors in economic policy and decision-making, particularly in the context of central bank communications. Woodford (2005) argues that increased central bank transparency and willingness to share assumptions about future policy have improved policy predictability and effectiveness. Boukus and Rosengerg (2006) showed that text analysis techniques such as Latent Semantic Analysis can provide valuable insights into market participants' reactions to FOMC minutes. Stephan Hansen's (2019) work on transparency in central bank communication is another example of the growing use of text analysis in economics. Hansen develops a new measure of central bank transparency based on NLP, which is used to assess the effectiveness of transparency policies in various countries. Overall, ongoing research in this area aims to produce even more accurate and efficient methods for analyzing financial text sentiment, which could have significant implications for financial decision-making and forecasting. More recently, doh et al. (2021) used natural language processing to identify the tone of FOMC post-meeting statements, highlighting the importance of qualitative statement language in policy decision-making. However, current linguistic analysis of FOMC minutes is still limited to vector language models and has yet to incorporate more advanced models such as transformers. Improving linguistic analysis techniques could provide even more accurate and efficient methods for analyzing central bank communications and informing economic policy.

## 3. Data

### 3.1. Findings

This study employs advanced language modeling techniques to randomly selected 1,065 sentences from the "Minutes of the Federal Open Market Committee" between January 3, 2006 and October 8, 2014, covering a total of 64 "Minutes" over the nine-year period. The study aims to gain insights into the language used by the

FOMC in their minutes and to explore the potential for enhancing language models. By utilizing advanced language modeling techniques, this study contributes to the importance of language and communication in economic text and highlights the potential for further advancements in language modeling techniques.

The analysis of the FOMC minutes yielded three key findings. Firstly, the FOMC follows a set of templates to cover economic situations, outlooks, and actions. Secondly, the FOMC is careful to avoid expressing emotion in their sentences, and they also recognize the impact of their statements on the public. In section 4, the study provides evidence of the neural sentiment of the FOMC minutes. Finally, the text of the minutes is highly condensed and contains a wealth of economic information, which presents a challenge for applying basic language models. To address this issue, the study proposes the use of "contextual matching" between nouns and verbs, adjectives and verbs, adverbs and nouns, and so on, to better capture the nuances of the language used. Neural network analyzers must also be careful to understand the economic rules employed by the FOMC.

### 3.2. Pre-processing

Existing research on sentiment analysis in FOMC minutes has typically utilized word-level analysis, even when dealing with longer paragraphs. However, this study seeks to explore the benefits of sentence-level sentiment analysis in order to gain a more contextual understanding of the Minutes. One of the initial challenges faced in this approach was the need to split the longer paragraphs into individual sentences, which required extensive pre-processing to ensure the data was properly cleaned and formatted for use in the language model.

The pre-processing steps used in this study included four main stages: splitting, removing stopwords, filtering for short-length sentences, and inspection. The first step involved splitting the longer paragraphs in the FOMC minutes into individual sentences, in order to facilitate sentence-level sentiment analysis. Next, we removed stopwords which do not carry significant meaning in the context of the sentences being analyzed. After removing stopwords, we filtered out sentences that were too short to provide meaningful insights, typically those with fewer than 8 words or 43chracters. Finally, we conducted a manual inspection of the remaining sentences to identify any errors or inaccuracies that may have resulted from the pre-processing steps. Overall, these pre-processing steps were critical in preparing the data for sentiment analysis and helped to ensure the accuracy and reliability of the results.

## 4. Method

### 4.1. VADER

VADER (Valence Aware Dictionary and sEntiment Reasoner) is a sentiment analysis tool that operates on a rule-based framework to evaluate the sentiment expressed in each text (Hutto and Gilbert, 2014). It is particularly effective in analyzing sentiments conveyed through social media posts, reviews, and forum discussions. VADER employs a lexicon of sentiment-related words and a set of rules to determine the polarity (positive, negative, or neutral) and intensity of sentiment in the text.

### 4.2. FinBERT

FinBERT is a pre-trained language model based on the BERT architecture (Devlin et al., 2019), which has been fine-tuned specifically for financial text data. The model is designed to capture the nuances and complexities of financial language and terminology, making it suitable for various financial text analysis tasks (Yang and Zhang, 2020). The model has been trained on a large corpus of financial text data, including news articles, corporate reports, and regulatory filings. In this study, a fine-tuned version of the FinBERT language model was utilized (Sarah-Yifei-Wang, 2021).

However, our analysis revealed that the FinBERT model did not perform perfectly when applied to FOMC minutes, as demonstrated in <Table 1>. For example, in sentence 2), which states that historically low levels of unemployment insurance benefits can be a positive signal for the economy, FinBERT classified the sentiment as 'Negative'. Moreover, the assigned sentiment score of -0.964 by FinBERT in sentence 2 is not only extreme but also in the opposite direction of the actual sentiment, which is concerning. This finding highlights a potential limitation of FinBERT in accurately capturing the sentiment of financial text data, especially in the context of FOMC minutes. Furthermore, compared to VADER, the FinBERT model seems to determine the sentiment in an excessively extreme manner and in some cases, in the wrong direction.

**<Table 1> Example of FinBERT Sentiment Analysis Result**

| Date | Sentence | Sentiment Classification | Score |
|---|---|---|---|
| 2008-02-20 | 1) On average, private nonfarm payroll employment in November and December rose at only about half of the average pace seen from July to October. | Negative (-1) | - 0.182 |
| 2019-11-20 | 2) The four-week moving average of initial claims for unemployment insurance benefits through mid-October remained near historically low levels. | Negative (-1) | -0.964 |
| 2021-08-18 | 3) Some other participants emphasized that recent high inflation readings had largely been driven by price increases in a handful of categories. | Neutral (0) | 0.076 |

The dataset used to train FinBERT includes Financial PhraseBank (Malo et al., 2014) and FiQA Task 1 sentiment scoring (Maia, 2018). While the model was trained on existing financial datasets, including a high agreement level dataset that was further fine-tuned by Sarah-Yifei-Wang (2021) to reduce noise in the labeled data, the dataset may not be well-suited for analyzing sentiment in FOMC minutes.

### 4.3. Labeling

To assess the precision of the language model, researchers meticulously annotated a total of 1,065 sentences. These annotated sentences were designated as the 'Actual sentiment'. During the labeling process, three key aspects of the FOMC minutes were considered. Given that the FOMC official records encompass diverse global economic scenarios, deriving a comprehensive sentiment can be challenging. Consequently, the researchers focused on 'Growth', 'Employment', and 'Inflation' as the main aspects. The outcomes of this labeling process are presented in <Table 2>.

**<Table 2> Example of Sentiment Labeling**

| Date | Sentence | Researcher Label (by aspect) | | | Language Model Label | |
|---|---|---|---|---|---|---|
| | | Growth | Employment | Inflation | VADER | FindBERT |
| 2008-08-26 | 1) Private nonfarm payroll employment fell in July at a pace only a bit less than the average monthly rate during the first six months of the year. | Neutral (0) | Negative (-1) | Neutral (0) | Neutral (0) | Negative (-1) |
| 2009-11-24 | 2) In the near term, most participants anticipated that substantial slack in both labor and product markets would likely keep inflation subdued. | Negative (-1) | Negative (-1) | Positive (1) | Negative (-1) | Negative (-1) |
| 2011-01-04 | 3) Exports of industrial supplies and agricultural goods registered the largest increases, although rising prices accounted for some of those gains. | Positive (1) | Neutral (0) | Negative (-1) | Positive (1) | Positive (1) |

In sentence 2), the phrase 'anticipated that substantial slack in both labor and product markets' conveys a negative sentiment regarding growth and employment. On the other hand, 'would likely keep inflation subdued' suggests a positive sentiment concerning inflation. However, both VADER and FinBERT analyze the sentence as having an overall negative sentiment. To compare the performance of VADER and FinBERT, we computed the average sentiment scores of three aspects. For sentence 2), the average score was -0.33, which was categorized as 'Negative', whereas for sentence 3), the average score was 0, indicating a 'Neutral' sentiment. These findings suggest that averaging the sentiment scores of multiple aspects can provide a more accurate assessment of the overall sentiment in financial text data, especially in cases where individual LM models may not perform perfectly.

# 5. Sentiment Prediction Results

## 5.1. Classification

As shown in <Figure 1>, the 'Actual' classification results, which were labeled by the researchers, exhibit a balanced distribution of sentiments. In contrast, both FinBERT and VADER demonstrated certain biases in their classifications. As previously discussed, FinBERT produced a slightly skewed, lopsided distribution, with a negative bias. However, VADER demonstrated an extremely centered distribution of sentiments, which prompted us to allocate a neutral sentiment rate of approximately 10% between -0.05 to 0.05. This adjustment was made to enhance the classification power of VADER and to better align with the 'Actual' sentiment distribution. After adjusting the neutral rate, VADER demonstrated a slightly skewed, positive sentiment distribution.

**<Figure1> Results of Sentiment Prediction**

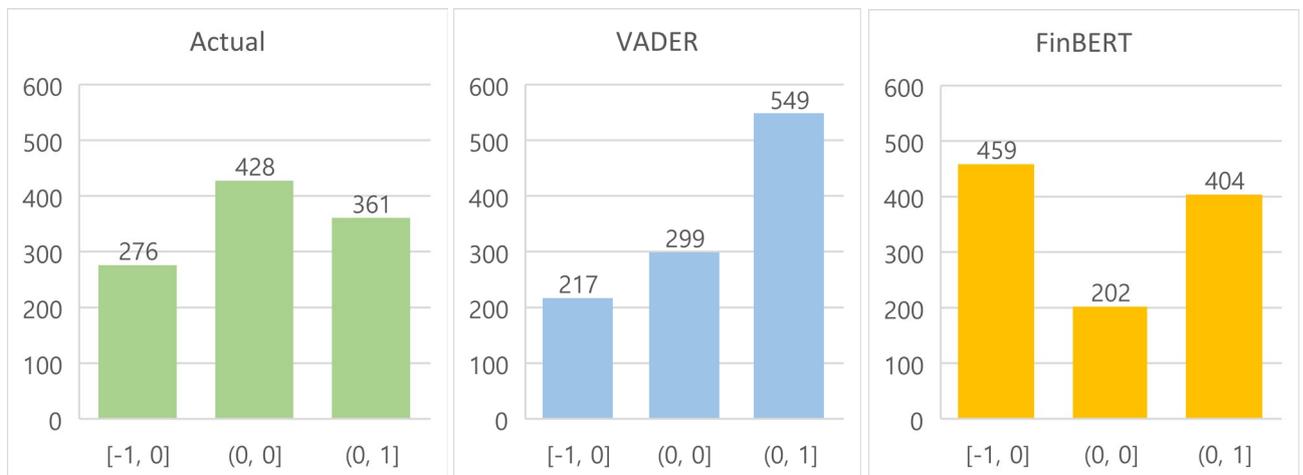

## 5.2. Confusion Matrix

The confusion matrix is a key evaluation metric used in sentiment analysis to measure the performance of a classification model. It is a matrix that summarizes the performance of a binary classifier by comparing the predicted labels with the true labels in the test dataset. The matrix consists of four elements: true positives (TP), true negatives (TN), false positives (FP), and false negatives (FN). In the context of sentiment analysis, the confusion matrix helps to evaluate the accuracy of the model in classifying positive, negative, and neutral sentiments. The true positives and true negatives represent the number of correctly classified samples for positive and negative sentiments, respectively. The false positives and false negatives represent the number of misclassified samples for positive and negative sentiments, respectively.

As shown in <Table 4>, the confusion matrix for VADER shows that the model achieved a relatively balanced performance in detecting positive, negative, and neutral sentiments. The model had the highest accuracy in detecting neutral sentiments, with 140 true positives and 88 true negatives, while the accuracy for positive and negative sentiments was lower. The model had a relatively high number of false positives and false negatives for both positive and negative sentiments. Overall, the VADER model achieved a moderate performance in sentiment analysis, but its accuracy could be improved for positive and negative sentiments.

The confusion matrix for FinBERT indicates that the model performed better in detecting positive and negative sentiments than neutral sentiments. The model had the highest accuracy in detecting negative sentiments, with 215 true negatives, and the accuracy for positive sentiments was also high, with 225 true positives. However, the model had a relatively low accuracy in detecting neutral sentiments, with 172 true negatives and 134 false positives. Overall, the FinBERT model achieved a better performance than VADER in detecting positive and negative sentiments, but its accuracy could be improved for neutral sentiments.

<Table 4> Multi-class Confusion Matrix Results

**VADER**

|  | Neutral (0) | Positive (1) | Negative (-1) |
|---|---|---|---|
| Neutral (0) | 140 | 226 | 62 |
| Positive (1) | 73 | 221 | 67 |
| Negative (-1) | 86 | 102 | 88 |

**FinBERT**

|  | Neutral (0) | Positive (1) | Negative (-1) |
|---|---|---|---|
| Neutral (0) | 172 | 134 | 122 |
| Positive (1) | 14 | 225 | 122 |
| Negative (-1) | 16 | 45 | 215 |

### 5.3. Receiver Operating Characteristic (ROC) Curve

In sentiment analysis, the ROC curve can be used to determine the optimal classification threshold for a model by analyzing the trade-off between the true positive rate and the false positive rate. A model that performs well in sentiment analysis should have a high true positive rate and a low false positive rate, indicating that it correctly identifies positive and negative sentiments while minimizing the misclassification of neutral sentiments. Furthermore, the area under the ROC curve (AUC) is a common metric used to quantify the overall performance of a sentiment analysis model. A higher AUC indicates better model performance in correctly identifying positive and negative sentiments.

Based on the AUC values provided in <Figure 2>, it appears that VADER's performance in identifying sentiment in financial text data is not particularly strong. In comparison to other sentiment analysis models, VADER may have some limitations in accurately detecting sentiment in financial text data, especially in cases where the sentiment is not strongly expressed. The AUC values for FinBERT suggest that the model performs relatively well in identifying negative sentiment in financial text data. An AUC value of 0.73 for the negative class indicates that the model has a high ability to distinguish negative sentiment from other classes. The AUC values for neutral and positive sentiment, both at 0.68, indicate a moderate level of performance in identifying those classes. Overall, these results suggest that FinBERT may be a useful tool for sentiment analysis in the financial domain, particularly for detecting negative sentiment in financial text data.

<Figure 2> ROC curve and AUC

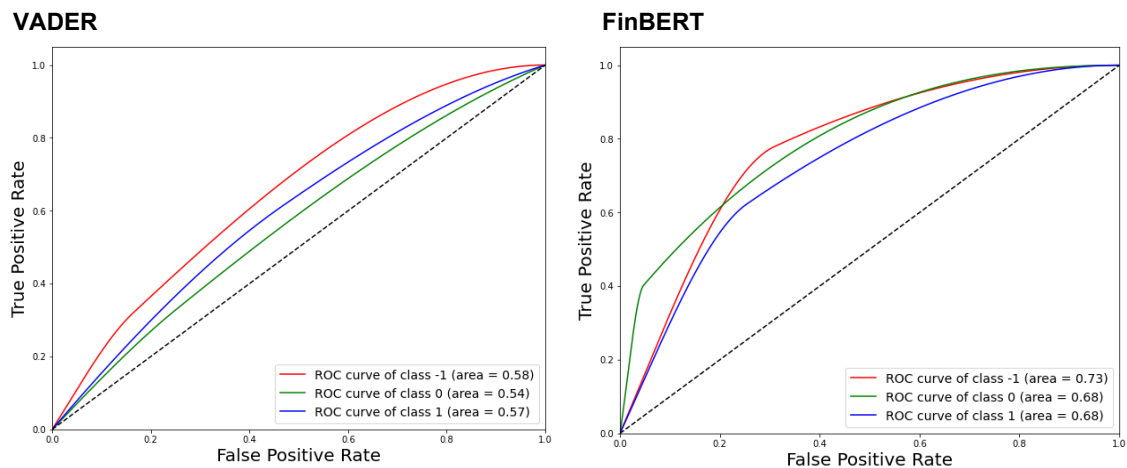

*note: class -1 = Negative / class 0 = Neutral / class 1 = Positive

Fewcett (2006) intoduced an AUC value of 0.7 or higher is generally considered good performance, while an AUC value below 0.5 indicates that the model is performing worse than random guessing. FinBERT performs relatively well in identifying negative sentiment in financial text data, with an AUC value above 0.7. The AUC values for neutral and positive sentiment are also moderately high, indicating a reasonable level of performance for these classes. Overall, based on the AUC criterion, FinBERT appears to outperform VADER in sentiment analysis of financial text data, particularly in identifying negative sentiment. However, it is important to note that AUC values are just one metric for evaluating model performance , and additional evaluation and comparison using other metrics and real-world applications may be needed for a more comprehensive assessment.

### 5.4. GPT-4 Prompt Test

This paper demonstrates that FinBERT is superior to other methods in extracting sentiment from FOMC data, while GPT's performance is comparatively weaker, despite being a cutting-edge technique. In this study, we employed GPT-4 to predict sentiment (negative/neutral/positive) for specific sentences, and the results are presented in <Table 5>. Our findings reveal that GPT-4 struggles to accurately discern sentiment in intricate sentences, with many being classified as 'Neutral' despite potentially being 'Positive' or 'Negative'. Therefore, we have included our interpretation and prediction of the sentiment for sentences where GPT-4 identified 'Neutral' sentiment. However, to make more precise sentiment predictions, the sentence must be considered in the context of the entire released statement, whereas in this study, the model predicted sentiment only for individual sentences. We also interpreted these sentences based on general concepts to determine their sentiment.

One of the biggest challenges of using GPT-4 is that it can be fine-tuned by researchers even if they don't fully understand how the model generates its sentiment predictions. This lack of interpretability is a significant limitation of large language models like GPT-4, which can make it difficult to trust and verify the accuracy of its output. Moreover, large language models require enormous amounts of computing power and data to train, which can be costly and time-consuming.

**<Table 5> Examples of GPT-4 Prediction Results**

| Date | Sentence | GPT4 | FinBERT | VADER |
|---|---|---|---|---|
| 2008-01-02 | "Debt in the domestic nonfinancial sector was estimated to be increasing somewhat more slowly in the fourth quarter than in the third quarter." | neutral | **positive** | **negative** |
| Interpretation by Author | *A gradual rise in debt indicates a stable market, and a slower pace of debt accumulation could be viewed as a favorable signal.* | **positive** | | |
| 2008-02-20 | "The trade-weighted foreign exchange value of the dollar against major currencies declined slightly, on balance, over the intermeeting period." | neutral | **negative** | **positive** |
| Interpretation by Author | *In general, the US dollar's value, calculated based on a weighted average of major foreign currencies, experienced a slight decline. This weakened value of the dollar could be perceived as a potentially negative indicator.* | **negative** | | |
| 2009-07-15 | "Meanwhile, the federal government issued large amounts of debt, and state and local government debt was estimated to have expanded moderately." | neutral | **positive** | **negative** |
| Interpretation by Author | *By taking on debt, the government can finance various services for the American public. Borrowing is generally considered a useful strategy for governments to fund crucial investments such as infrastructure and education, among other areas.* | **positive** | | |
| 2009-11-24 | "In the near term, most participants anticipated that substantial slack in both labor and product markets would likely keep inflation subdued." | neutral | **negative** | **positive** |
| Interpretation by Author | *It is generally desirable to maintain a stable level of inflation that is neither too high nor too low. In this context, the statement may be seen as positive for the economy* | **positive** | | |
| 2013-07-10 | "However, some others commented that any adverse effects of the increase in rates on financial conditions more broadly appeared to be limited." | neutral | **positive** | **negative** |
| Interpretation by Author | *According to these individuals, the adverse effects of increasing interest rates on the overall financial situation were either insignificant or not severe.* | **positive** | | |
| 2014-01-08 | "Year-to-date issuance of commercial mortgage-backed securities (CMBS) remained strong, but far below levels seen before the financial crisis." | neutral | **positive** | **negative** |
| Interpretation by Author | *CMBS plays a role in providing liquidity to the market. Although the issuance of CMBS has been robust this year, it remains substantially below the pre-2008 financial crisis levels.* | **negative** | | |

## 6. Conclusion and Future work

In conclusion, our analysis shows that the unique characteristics of the FOMC minutes, such as complex sentence structures and lower emotional content, make them challenging for current NLP techniques to analyze. However, our study indicates that FinBERT outperforms other techniques in accurately predicting negative sentiment.

Nonetheless, further improvements are necessary to optimize the performance of both FinBERT and GPT-4 in analyzing FOMC sentiment. To overcome the limitations of existing NLP techniques, alternative approaches such as combining NLP techniques with statistical models or expert opinions should be explored.

In summary, our study highlights the challenges and limitations of using current NLP techniques to analyze FOMC texts. While FinBERT shows promise in predicting negative sentiment accurately, additional enhancements are necessary for both FinBERT and GPT-4. Future research should focus on improving sentiment analysis accuracy for FOMC texts and exploring alternative approaches to gain more comprehensive economic insights.